\RequirePackage[T1]{fontenc}
\documentclass[letterpaper,10pt,conference,nofonttune]{ieeeconf}
\IEEEoverridecommandlockouts
\usepackage{newtxtext}
\usepackage{amsmath,amssymb}
\usepackage{newtxmath}
\usepackage{booktabs}
\usepackage{graphicx}
\usepackage{multirow}
\usepackage{cite}
\usepackage{url}

\newif\ificraanonymous
\icraanonymousfalse
\newcommand{\micropush}{\textsc{MicroPush}}
\newcommand{\um}{\ensuremath{\mu\mathrm{m}}}
\newcommand{\ums}{\ensuremath{\mu\mathrm{m}\,\mathrm{s}^{-1}}}

\title{\LARGE \bf
MicroPush: A Lightweight Simulator for Sim-to-Real Microrobotic Pushing and Assembly
}

\ificraanonymous
\author{Anonymous Authors}
\else
\author{Yanda Yang, Zion Lee, and Sambeeta Das%
\thanks{Y. Yang, Z. Lee, and S. Das are with the Department of Mechanical Engineering,
University of Delaware, Newark, DE 19716, USA
(e-mail: robotyyd@udel.edu; zionlee@udel.edu; samdas@udel.edu).}}
\fi

\begin{document}
\maketitle
\thispagestyle{empty}
\pagestyle{empty}

\begin{abstract}
Developing learning-based autonomy for microrobotic manipulation is difficult because physical trials are slow, variable, and poorly suited to large-scale training. We present \micropush{}, a lightweight simulation and benchmarking environment for pushing and assembly with magnetic rolling microrobots. The simulator combines overdamped planar dynamics with calibrated rolling actuation, reduced-order contact handling, short-range interaction corrections, and optional background flow, while providing common interfaces for planning, model-based control, reinforcement learning (RL), interactive visualization, and headless evaluation. In no-render tests, \micropush{} remains substantially faster than real time, reaching $37.0\times$ real time with 100 passive particles. We further evaluate the simulator through zero-shot sim-to-real policy transfer. A Residual RL--MPC policy is trained entirely in \micropush{}, frozen after simulation training, and deployed on the physical system without real-world policy fine-tuning. A $20$~\um{} magnetic microrobot pushes a similarly sized silica particle along a figure-eight trajectory with $1.91$~\um{} median and $4.65$~\um{} 95th-percentile path error, and the same autonomy stack performs sequential line, triangle, and six-particle hexagonal assembly. These results show that a deliberately reduced-order simulator can provide an efficient training and development substrate for sim-to-real microrobotic autonomy.
\end{abstract}

\section{Introduction}

Magnetic microrobots enable wireless manipulation without onboard power and have been widely explored for transport, positioning, and assembly of biological and synthetic objects in confined microscopic workspaces \cite{wang2023untethered,zhang2019robotic,yang2024quadrupole}. Surface-rolling magnetic microrobots are particularly attractive for near-surface manipulation because their motion can be controlled through externally applied rotating magnetic fields while maintaining a compact, untethered form factor \cite{pieters2015rodbot,nguyen2021magnetically,yang2023rolling}. Moving from teleoperated demonstrations to autonomous manipulation requires perception, planning, and feedback control to remain reliable under intermittent contact, localization error, surface interactions, and fluid disturbances \cite{pawashe2011two,feemster2020autonomous,sokolich2025autonomous,yang2023closed}.

Microrobotic pushing is especially challenging because viscous effects dominate inertia and near-wall hydrodynamics, friction, adhesion, and contact loss can strongly affect object motion \cite{bozuyuk2022reduced,mohanty2022ceflowbot,dillinger2021ultrasound}. Physical experimentation is also difficult to scale: repeated trials are time consuming, experimental conditions vary, and large controller sweeps or reinforcement-learning data collection are impractical on hardware alone \cite{jitosho2021dynamics,huang2020dynamic,tan2025interactive}. These challenges motivate simulation environments for repeated autonomy development and large-scale policy training before physical deployment.

Simulation and standardized benchmarks have accelerated robot learning and manipulation by enabling rapid iteration, controlled ablations, and reproducible comparison \cite{todorov2012mujoco,zhu2020robosuite,james2020rlbench,makoviychuk2021isaac}. Microrobotics-specific environments have also emerged for different actuation modalities \cite{amokrane2018macro,tong2025robotic,tan2025interactive,wang2026magrobot}. However, computationally efficient tools remain limited for surface-rolling magnetic microrobots performing contact pushing and sequential assembly under clutter and flow. General-purpose simulators do not directly capture the operating assumptions of viscous-dominated microscale manipulation, while high-fidelity fluid--structure simulation is costly for large-scale closed-loop learning and benchmark evaluation.

Autonomous microrobotic manipulation has been demonstrated through magnetic contact pushing \cite{pawashe2011two,feemster2020autonomous}, non-contact manipulation \cite{jia2024efficient,jia2025non,sokolich2025autonomous}, and dedicated pushing controllers \cite{koua2026pushing}. These studies demonstrate the feasibility of autonomous microscale manipulation, while a unified environment for simulation-based development, planning, learning, and sequential multi-object evaluation remains lacking.

We present \micropush{}, a lightweight simulation and benchmarking environment for microrobotic pushing and assembly. \micropush{} combines calibrated rolling actuation with overdamped planar dynamics, reduced-order interaction handling, optional background flow, global and task-level planning, model-based control, a Gym-compatible learning interface \cite{brockman2016openai}, interactive visualization, and headless evaluation. Beyond simulation-only evaluation, we use the Residual RL--MPC controller introduced in \cite{yang2026resrl} to test direct policy transfer: the residual policy is trained entirely in \micropush{}, frozen after training, and deployed on a physical magnetic microrobot without real-world policy fine-tuning. Experiments include continuous figure-eight pushing and sequential line, triangle, and six-particle hexagonal assembly.

The main contributions are:
\begin{enumerate}
\item a lightweight microrobotic manipulation environment integrating calibrated rolling actuation, reduced-order interaction dynamics, background flow, planning, feedback control, reinforcement learning, visualization, and headless evaluation;
\item a reproducible benchmark suite for contact transport and sequential multi-object assembly, with substantially faster-than-real-time simulation for training and systematic evaluation; and
\item zero-shot sim-to-real validation of a Residual RL--MPC policy trained entirely in \micropush{} for continuous pushing and sequential multi-particle assembly.
\end{enumerate}

\begin{figure*}
    \centering
    \includegraphics[width=0.8\textwidth]{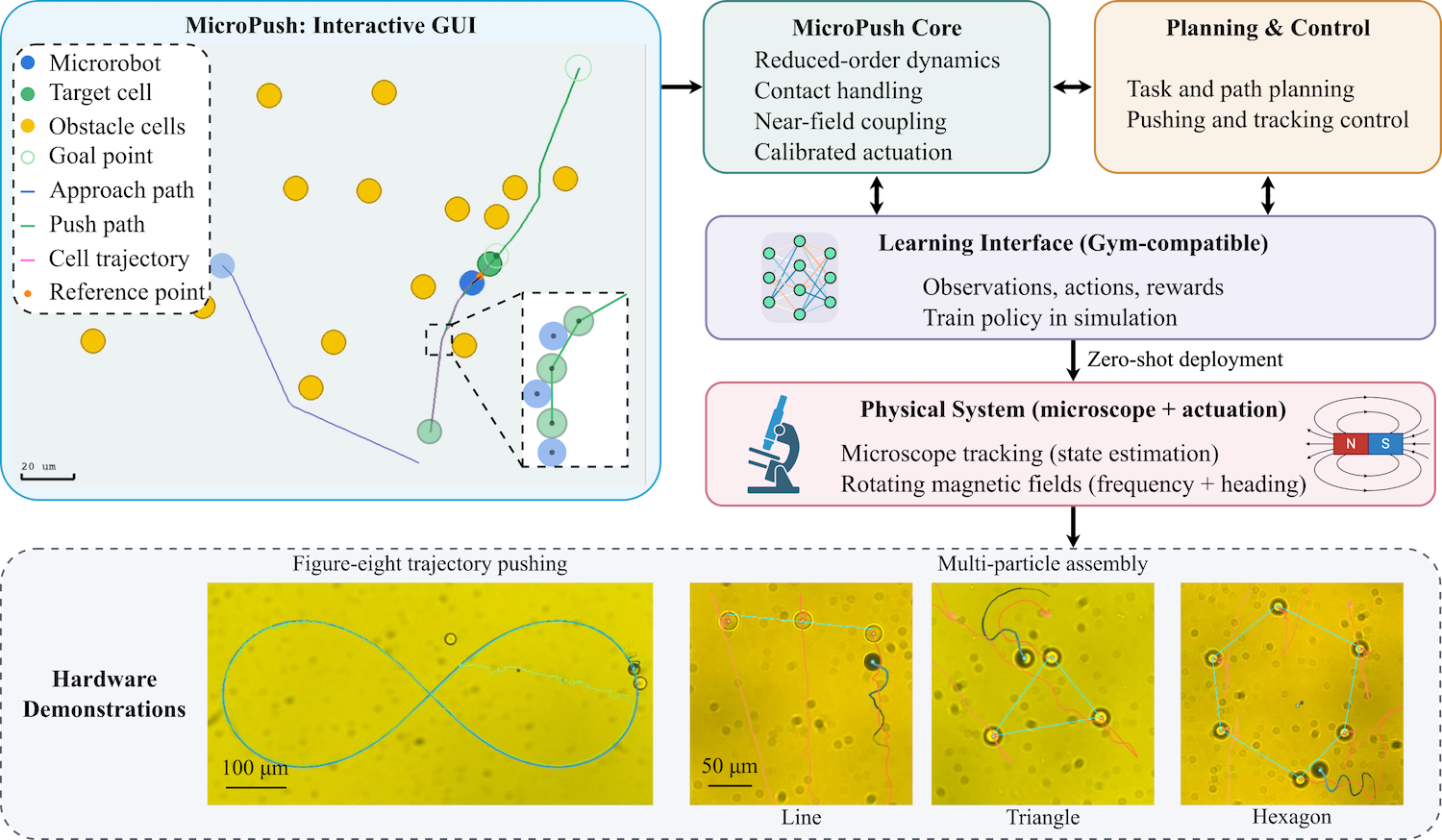}
    \caption{\micropush{} architecture and sim-to-real workflow. The interactive GUI and simulator core expose reduced-order interaction dynamics, calibrated actuation, planning/control modules, and a Gym-compatible learning interface. Policies are trained in simulation and deployed through microscope-based state estimation and rotating-field actuation. The bottom row previews the two physical evaluations used in this paper: continuous figure-eight pushing and sequential multi-particle assembly.}
    \label{fig:system}
\end{figure*}

\section{MicroPush Environment}
\subsection{System architecture and interfaces}
\micropush{} is organized around a shared simulator core accessed by the interactive GUI, planning and control modules, and a Gym-compatible learning interface (Fig.~\ref{fig:system}). At each step, an algorithm receives robot/object states and returns an actuation command $a_t=(\omega_t,\theta_t)$, where $\omega_t$ is the rotating-field frequency and $\theta_t$ is the desired rolling direction. The same planar command abstraction is used by classical controllers and learned policies. During physical deployment, simulated observations are replaced by microscope-based state estimates and the command is routed to the magnetic actuation system, minimizing changes to the autonomy stack between training and experiment.

The environment supports deterministic seeding, scenario generation, trajectory logging, and headless execution. Global planning generates collision-aware transfer paths, task-level planning assigns and sequences multiple objects, and local control executes staged contact acquisition and pushing. This separation allows the same simulator to support interactive debugging, reproducible benchmark evaluation, and large-scale policy training.

\subsection{Calibrated actuation and overdamped motion}
Let $\mathbf p_i,\mathbf v_i\in\mathbb{R}^2$ denote the planar position and velocity of body $i$. At the microscale, inertia is neglected and motion is represented in overdamped form,
\begin{equation}
\mathbf v_i
=
\mathbf u_i
+
\gamma_i^{-1}
\left(
\mathbf F_i^{\rm int}
+
\mathbf F_i^{\rm wall}
\right)
+
\mathbf v^{\rm flow}(\mathbf p_i),
\label{eq:overdamped}
\end{equation}
where $\gamma_i$ is an effective drag coefficient, $\mathbf F_i^{\rm int}$ denotes effective pairwise interaction terms in simulator units, $\mathbf F_i^{\rm wall}$ enforces the workspace boundary, and $\mathbf v^{\rm flow}$ is an optional background-flow velocity. Passive particles have $\mathbf u_i=\mathbf 0$. For the microrobot,
\begin{equation}
\mathbf u_r
=
k_v\omega
\begin{bmatrix}
\cos\theta\\
\sin\theta
\end{bmatrix},
\qquad
0\leq\omega\leq\omega_{\max},
\label{eq:actuation}
\end{equation}
where $\omega$ is the rolling frequency and $\theta$ is the commanded heading.

\begin{figure}
    \centering
    \includegraphics[width=0.8\linewidth]{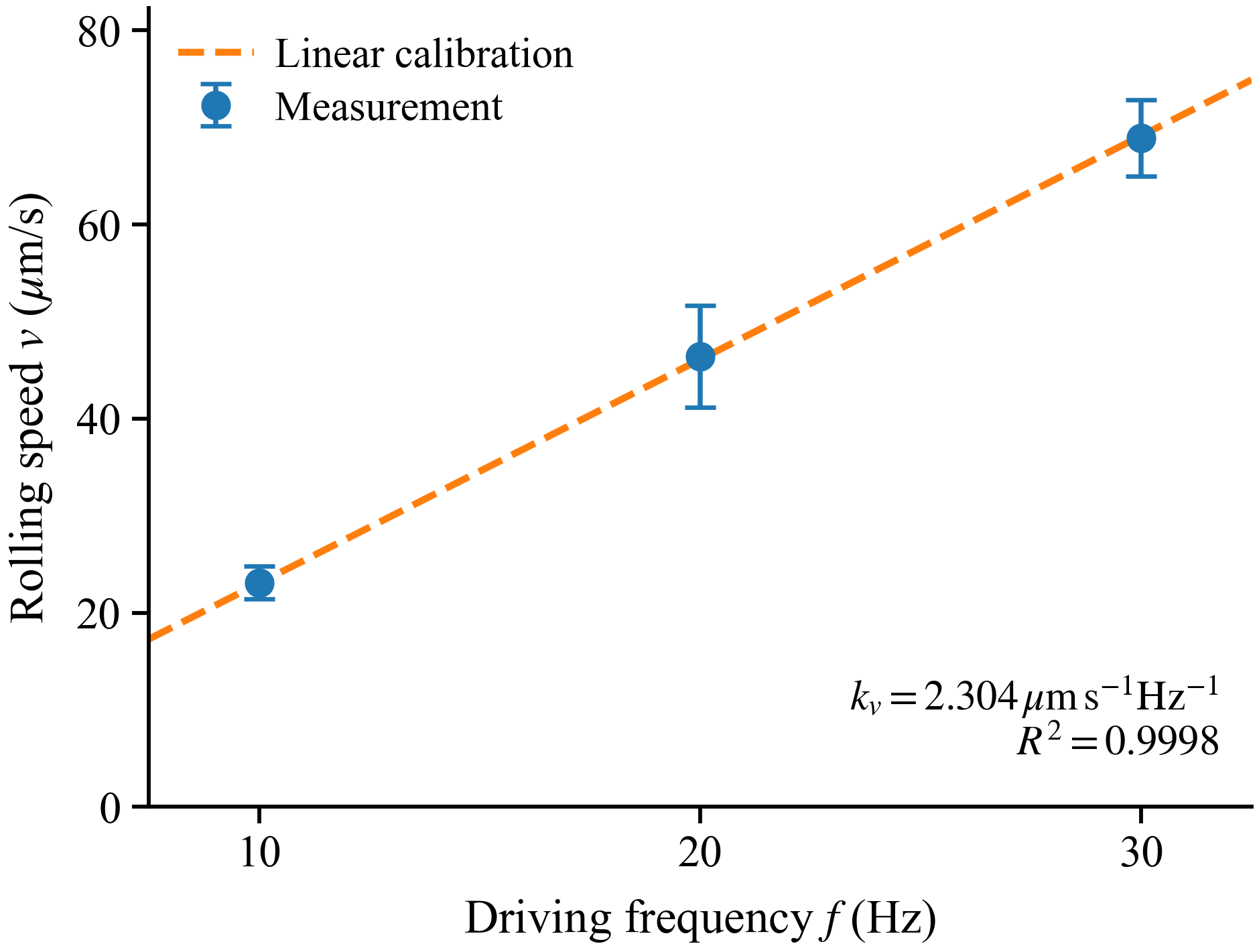}
    \caption{Free-space rolling-speed calibration used by \micropush{}. Markers show measured mean speed with error bars; the dashed line is the through-origin linear fit used in Eq.~\eqref{eq:actuation}.}
    \label{fig:speed}
\end{figure}

The coefficient $k_v$ is identified experimentally from free rolling. As shown in Fig.~\ref{fig:speed}, the measured speed is approximately linear over the calibrated low-frequency operating range, giving $k_v=2.304~\um\,\mathrm{s}^{-1}\mathrm{Hz}^{-1}$ ($R^2=0.9998$). We therefore restrict control to this range and use Eq.~\eqref{eq:actuation} as the common simulator/hardware actuation map. Bounded perturbations of speed magnitude and heading are applied during simulation to represent actuation variability. Positions are advanced with explicit Euler integration,
\begin{equation}
\mathbf p_i^{\,t+1}
=
\mathbf p_i^{\,t}
+
\Delta t\,\mathbf v_i^{\,t}.
\label{eq:euler}
\end{equation}

\subsection{Contact handling and empirical near-field coupling}
\begin{figure}
    \centering
    \includegraphics[width=1.0\linewidth]{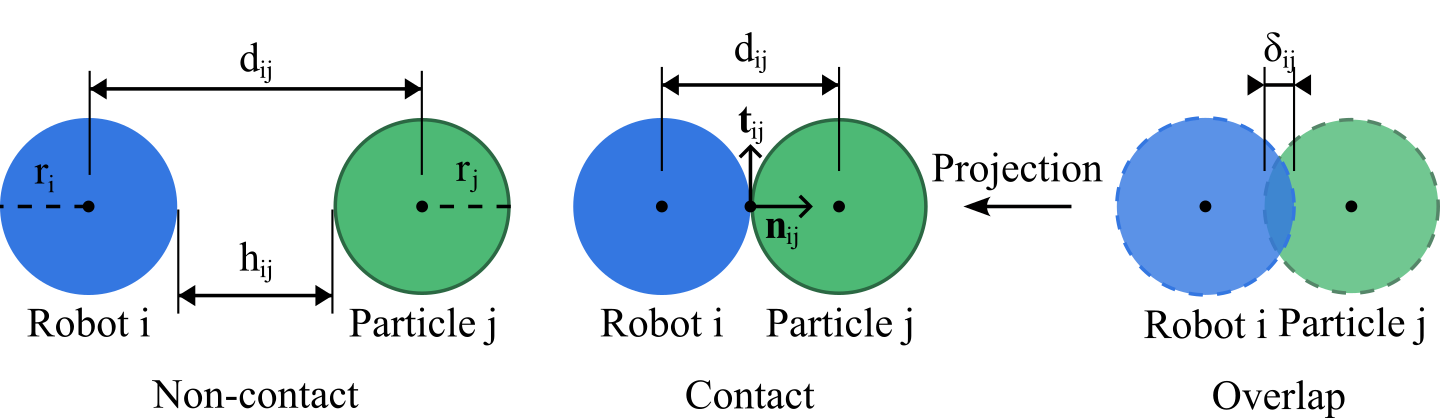}
    \caption{Pairwise disk geometry used by the reduced-order interaction model. In non-contact, $h_{ij}$ is the surface gap. At contact, relative motion is resolved along the unit normal $\mathbf n_{ij}$ and tangent $\mathbf t_{ij}$; nonpenetration is ultimately enforced by the position projection in Eq.~\eqref{eq:projection}.}
    \label{fig:contact}
\end{figure}

Each robot or manipulated object is represented by a disk. For a pair $(i,j)$, define the center distance, surface gap, and unit normal as
\begin{equation}
d_{ij}=\|\mathbf p_j-\mathbf p_i\|_2,\qquad
h_{ij}=d_{ij}-(r_i+r_j),\qquad
\mathbf n_{ij}=\frac{\mathbf p_j-\mathbf p_i}{d_{ij}},
\label{eq:geometry}
\end{equation}
where $\mathbf n_{ij}$ points from body $i$ to body $j$. The corresponding unit tangent $\mathbf t_{ij}$ is orthogonal to $\mathbf n_{ij}$, and the overlap is
\begin{equation}
\delta_{ij}=\max(0,-h_{ij}).
\label{eq:overlap}
\end{equation}
Figure~\ref{fig:contact} summarizes these quantities.

For overlapping pairs, the implementation computes a nonlinear overlap-response magnitude
\begin{equation}
\eta_{ij}^{n}=k_h\delta_{ij}^{3/2},\qquad \delta_{ij}>0,
\label{eq:overlap_response}
\end{equation}
inside the pairwise velocity update. After integration, nonpenetration is enforced explicitly: if the post-update overlap is $\delta_{ij}^{\rm post}>0$, the centers are projected apart along $\mathbf n_{ij}$,
\begin{equation}
\mathbf p_i\leftarrow\mathbf p_i-\frac{1}{2}\delta_{ij}^{\rm post}\mathbf n_{ij},\qquad
\mathbf p_j\leftarrow\mathbf p_j+\frac{1}{2}\delta_{ij}^{\rm post}\mathbf n_{ij},
\label{eq:projection}
\end{equation}
and any remaining approaching normal component of the pairwise relative velocity is removed. A position-level projection is applied after integration to remove residual overlap, ensuring a nonpenetrating configuration at the end of each simulation step.

Tangential relative motion is handled with a stick--slip update. With $\mathbf v_{ij}^{\rm rel}=\mathbf v_i-\mathbf v_j$ and $v_{ij}^{t}=\mathbf v_{ij}^{\rm rel}\!\cdot\!\mathbf t_{ij}$, the available per-step tangential correction is limited by
\begin{equation}
\Delta v_{ij,t}^{\max}
=
\mu\eta_{ij}^{n}
\left(\gamma_i^{-1}+\gamma_j^{-1}\right)\Delta t,
\label{eq:friction}
\end{equation}
where $\mu$ is an effective tangential parameter. The applied tangential correction cancels the relative tangential motion when
$|v^t_{ij}|\leq \Delta v^{\max}_{ij,t}$ (sticking), and otherwise reduces its
magnitude by $\Delta v^{\max}_{ij,t}$ while preserving its direction (slipping).

The positive-gap interaction is handled separately. Let $\dot h_{ij}=(\mathbf v_j-\mathbf v_i)\!\cdot\!\mathbf n_{ij}$. The implementation used for the reported results applies two numerical corrections: a narrow guard-band update during approach and a separate short-range update during separation,
\begin{equation}
\dot h_{ij}^{\rm corr}=\begin{cases}
2\dot h_{ij}, & 0<h_{ij}\leq g,\;\dot h_{ij}<0,\\
(1-\beta)\dot h_{ij}, & 0<h_{ij}\leq h_{\rm th},\;\dot h_{ij}>0,\\
\dot h_{ij}, & \text{otherwise},
\end{cases}
\label{eq:nearfield_piecewise}
\end{equation}
with
\begin{equation}
h_{\rm th}=\operatorname{clip}\!\left(h_0+\alpha\max(0,\dot h_{ij}),\,h_0,\,h_{\max}\right).
\label{eq:nearfield_range}
\end{equation}
The first branch advances an approaching pair toward contact inside a small numerical guard band, whereas the second slows separation within the configured near-field range. We therefore treat these rules as an empirical near-contact coupling heuristic, not as a calibrated lubrication model.

When background flow is enabled, \micropush{} adds an idealized planar Poiseuille profile,
\begin{equation}
\mathbf v^{\rm flow}(y)=
\begin{bmatrix}
U_{\max}\left[1-\left(2y/H-1\right)^2\right]\\
0
\end{bmatrix},
\label{eq:flow}
\end{equation}
where $H$ is the channel height and $U_{\max}$ is the centerline flow speed. This profile provides a controllable disturbance for evaluating contact loss and recovery. Overall, the model remains deliberately reduced order: rigid planar objects, empirical near-contact coupling, idealized flow, and numerical nonpenetration are used in place of deformation, three-dimensional fluid--structure interaction, or force-level contact prediction. The intended use is fast closed-loop learning and algorithm comparison.

Representative simulator interactions are provided in Movies S1--S4.

\section{Planning, Control, and Learning}
\subsection{Collision-aware global and task planning}
\begin{figure}
    \centering
    \includegraphics[width=\linewidth]{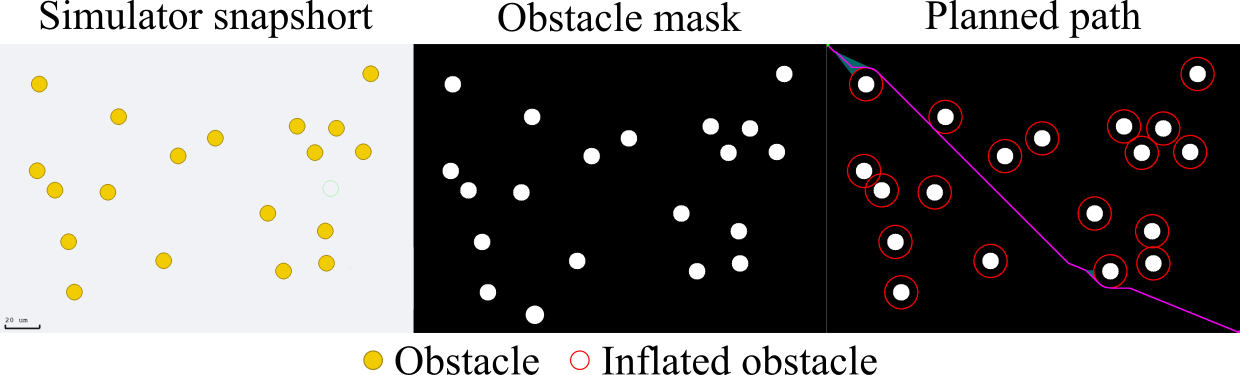}
    \caption{Global-planning interface. A simulator snapshot is converted to an obstacle mask, obstacles are inflated according to Eq.~\eqref{eq:inflate}, and a collision-aware path is generated for the moving body.}
    \label{fig:global}
\end{figure}

The global planning interface converts the current workspace into an occupancy representation and accounts for the finite radius of the moving body through obstacle inflation. For obstacle $k$ with radius $r_k^{\rm obs}$, the effective planning radius is
\begin{equation}
R_k=r_k^{\rm obs}+r_{\rm move}+r_{\rm pad},
\label{eq:inflate}
\end{equation}
where $r_{\rm move}$ is the radius of the moving body and $r_{\rm pad}$ is an optional safety margin. Figure~\ref{fig:global} illustrates the simulator scene, inflated obstacle representation, and resulting collision-aware path. Different global planners can consume the same inflated scene representation and return a resampled waypoint sequence to downstream control.

For multi-object assembly, target geometry, particle--target assignment, and transfer ordering are handled above the global planner. A regular $K$-vertex target centered at $\mathbf c\in\mathbb{R}^2$ is defined by
\begin{equation}
\mathbf q_j
=
\mathbf c
+
\rho
\begin{bmatrix}
\cos(2\pi j/K+\phi)\\
\sin(2\pi j/K+\phi)
\end{bmatrix},
\qquad j=0,\ldots,K-1,
\label{eq:vertices}
\end{equation}
where $\rho$ is the formation radius and $\phi$ specifies its angular orientation.

\begin{figure}
    \centering
    \includegraphics[width=\linewidth]{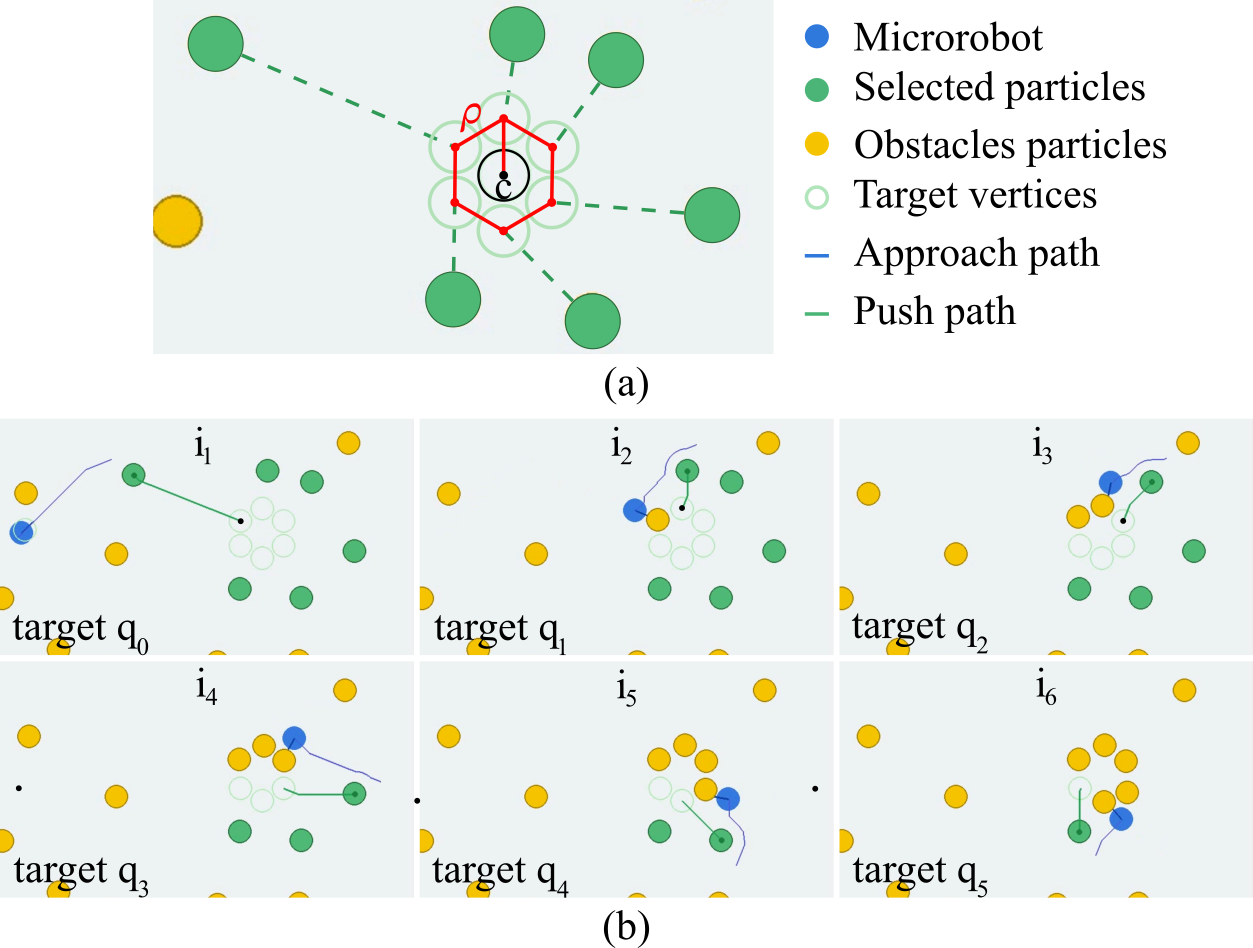}
    \caption{Task-level planning for six-particle hexagonal assembly. Top: target generation and particle--target assignment. Bottom: sequential transfers; for each selected particle, the global planner generates an approach path for contact acquisition and a push path toward the currently assigned target. See Movie S5 for the corresponding assembly execution.}
    \label{fig:task}
\end{figure}

For the hexagonal task, $K=6$. An initial particle--target assignment is obtained by minimizing total Euclidean assignment cost with the Hungarian algorithm. The resulting transfers are ordered using a greedy nearest-neighbor heuristic with 2-opt refinement. A lightweight local search then swaps particle--target assignments when doing so reduces the combined placement and inter-transfer travel cost. Each transfer is replanned from the current scene and decomposed into an approach path for the microrobot and a push path for the selected particle (Fig.~\ref{fig:task}).

\subsection{Contact-aware pushing control}
\begin{figure*}
    \centering
    \includegraphics[width=0.8\textwidth]{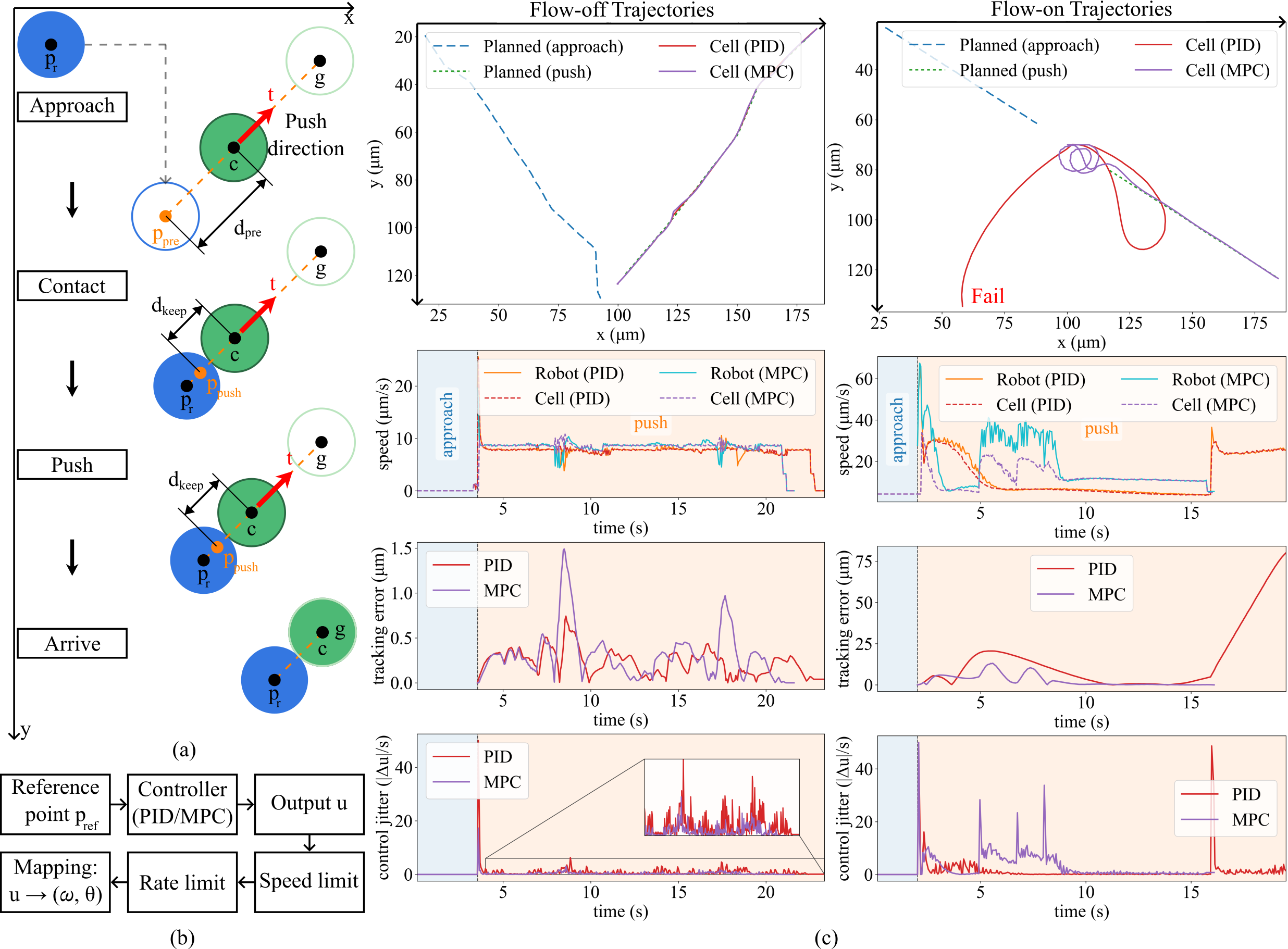}
    \caption{Contact-aware pushing and representative simulator executions. (a) Staged approach, contact, push, and arrival references relative to the particle and desired transfer direction. (b) Shared low-level command pipeline: the PID/MPC output is subjected to speed and rate limits and mapped to rolling frequency and heading. (c) Representative closed-loop rollouts without flow (left) and with background flow (right), including trajectories, speeds, tracking error, and command variation. In the shown flow-on trial, PID loses the particle whereas MPC maintains the transfer. Shaded intervals distinguish approach and push phases. See Movie S6 for the corresponding flow-on comparison.}
    \label{fig:control}
\end{figure*}

Directly driving the microrobot toward a particle waypoint can establish contact from an unfavorable direction. Let $\mathbf p_r$ and $\mathbf c$ denote the robot and particle centers and let $\mathbf g$ be the next particle waypoint. The desired pushing direction is
\begin{equation}
\mathbf t=\frac{\mathbf g-\mathbf c}{\|\mathbf g-\mathbf c\|_2}.
\label{eq:push_direction}
\end{equation}
Before contact, \micropush{} guides the robot to a pre-contact reference behind the particle,
\begin{equation}
\mathbf p_{\rm pre}=\mathbf c-d_{\rm pre}\mathbf t,
\qquad d_{\rm pre}=r_r+r_o+d_0,
\label{eq:precontact}
\end{equation}
where $r_r$ and $r_o$ are the robot and particle radii and $d_0$ is an additional approach offset. Once contact is established, the controller switches to
\begin{equation}
\mathbf p_{\rm push}=\mathbf c-d_{\rm keep}\mathbf t,
\qquad d_{\rm keep}<d_{\rm pre},
\label{eq:push_reference}
\end{equation}
which keeps the robot close to the particle during pushing. The reference is updated as the particle advances, and the transfer terminates when the particle reaches the goal region around $\mathbf g$. Figure~\ref{fig:control}(a) summarizes the approach--contact--push--arrival sequence.

The nominal MPC uses the planar integrator model
\begin{equation}
\mathbf x_{k+1}=\mathbf x_k+\Delta t\,\mathbf u_k
\label{eq:mpc_dynamics}
\end{equation}
and solves
\begin{align}
\min_{\mathbf u_{0:N-1}}\quad
&\sum_{k=0}^{N-1}
\left(
\|\mathbf x_{k+1}-\mathbf p_{\rm ref}\|_{Q}^{2}
+
\|\mathbf u_k\|_{R}^{2}
\right)
\nonumber\\
&+\|\mathbf x_N-\mathbf p_{\rm ref}\|_{Q_f}^{2}
+\|\mathbf u_0-\mathbf u_{-1}\|_{S}^{2},
\label{eq:mpc}
\end{align}
where $\|\mathbf z\|_Q^2\equiv\mathbf z^TQ\mathbf z$. Only the first optimized command is applied in receding-horizon fashion. Both MPC and PID use the same contact-aware reference and the same speed/rate envelope,
\begin{equation}
\|\mathbf u_k\|_2\leq v_{\max},\qquad
\|\mathbf u_k-\mathbf u_{k-1}\|_2\leq\Delta u_{\max},
\label{eq:control_limits}
\end{equation}
and the resulting planar velocity is mapped to rolling frequency and heading through Eq.~\eqref{eq:actuation}. Figure~\ref{fig:control}(b) summarizes this shared command pipeline.

Figure~\ref{fig:control}(c) compares representative flow-off and flow-on executions. Both controllers track similarly without flow, whereas the disturbed trial exposes contact loss and large tracking error for PID while MPC maintains pushing and reaches the goal. This qualitative comparison motivates the controlled benchmark in Sec.~\ref{sec:sim_eval}.

\section{Simulation Evaluation}
\label{sec:sim_eval}
\subsection{Benchmark tasks and protocol}
We use two canonical tasks. Transport requires one selected object to be moved to a goal in clutter, with flow either disabled or enabled. Assembly assigns six objects to a regular hexagonal formation and executes the transfers sequentially. Seeded initial conditions, timeouts, actuation limits, and success definitions are fixed within each benchmark condition, and both successful and failed episodes are retained. The closed-loop benchmark uses the settings in Table~\ref{tab:settings}; continuous metrics are summarized only over successful trials.

\begin{table}
\caption{Key settings for the closed-loop simulation benchmark.}
\label{tab:settings}
\centering
\footnotesize
\setlength{\tabcolsep}{4pt}
\begin{tabular}{@{}lc@{}}
\toprule
Setting & Value\\
\midrule
Simulation timestep $\Delta t$ & 0.05 s\\
Robot / particle diameter & 10.0 / 10.0 $\um$\\
Maximum rolling frequency $\omega_{\max}$ & 30 Hz\\
Speed / heading perturbation & $\pm2\%$ / $\pm0.02$ rad\\
Flow-on centerline speed $U_{\max}$ & 5.0 $\um$/s\\
MPC horizon & 10\\
Success radius & 0.6 $\um$\\
\bottomrule
\end{tabular}
\end{table}

\subsection{Controller benchmark results}
For the results in Table~\ref{tab:sim}, success is computed over all episodes. Among successful episodes, $E_{\Delta\omega}=\sum_t|\omega_t-\omega_{t-1}|$ measures command variation, and tracking error is additionally reported for flow-off transport. Under flow-off transport, both controllers achieve near-perfect success, with MPC yielding slightly lower tracking error and command variation. With background flow enabled, MPC maintains 100\% success whereas PID drops to 41.2\%. In six-particle assembly, MPC also achieves higher success (86.7\% versus 66.7\%) and lower command variation.

\begin{table}
\caption{\micropush{} controller benchmark; success is computed over all episodes; continuous metrics are medians over successful episodes.}
\label{tab:sim}
\centering
\footnotesize
\setlength{\tabcolsep}{3.2pt}
\begin{tabular}{@{}llccccc@{}}
\toprule
Task & Flow & Ctrl. & $N$ & Success & Track ($\um$) & $E_{\Delta\omega}$\\
\midrule
Transport & on  & MPC & 80 & 100\%  & --    & 353\\
Transport & on  & PID & 80 & 41.2\% & --    & 418\\
\midrule
Transport & off & MPC & 80 & 98.8\% & 0.159 & 216\\
Transport & off & PID & 80 & 100\%  & 0.189 & 250\\
\midrule
Assembly  & off & MPC & 30 & 86.7\% & --    & 690\\
Assembly  & off & PID & 30 & 66.7\% & --    & 841\\
\bottomrule
\end{tabular}
\end{table}

\subsection{Computational efficiency}
Fast execution is important for RL training and large-scale evaluation.
We therefore measured no-render simulator throughput for scenes containing
1, 10, 50, and 100 passive particles, with one additional microrobot in
each scene. Each configuration was simulated for 1,000 steps and repeated
five times. The real-time factor (RTF) is defined as
\begin{equation}
\mathrm{RTF}
=
\frac{T_{\rm sim}}{T_{\rm wall}}
=
f_{\rm step}\Delta t,
\label{eq:rtf}
\end{equation}
where $f_{\rm step}$ is the measured simulator throughput in steps/s and
$\Delta t=0.05$~s. Thus, $\mathrm{RTF}>1$ indicates faster-than-real-time
execution.

As shown in Table~\ref{tab:throughput}, throughput decreases as the number
of interacting particles increases, but remains substantially faster than
real time over all tested scene sizes. Even with 100 passive particles,
MicroPush achieves an average throughput of approximately 740 steps/s,
corresponding to $37.0\times$ real time. These results support efficient
policy training and large-scale simulation-based evaluation.

\begin{table}
\caption{No-render throughput with $\Delta t=0.05$~s over five runs. Steps/s is mean $\pm$ SD.}
\label{tab:throughput}
\centering
\footnotesize
\begin{tabular}{@{}rrr@{}}
\toprule
Passive particles & Steps/s (mean $\pm$ SD) & Mean RTF\\
\midrule
1   & $253176 \pm 10159$ & $12659\times$\\
10  & $50164 \pm 929$    & $2508\times$\\
50  & $2934 \pm 27$      & $146.7\times$\\
100 & $740 \pm 8$        & $37.0\times$\\
\bottomrule
\end{tabular}
\end{table}

\section{Zero-Shot Sim-to-Real Experiments}

\subsection{Residual RL--MPC Policy}

The Gym-compatible interface exposes the simulated dynamics, contact states,
disturbances, and actuation limits to learned policies. To evaluate whether
\micropush{} can support policy learning beyond simulation-only benchmarks,
we use the Residual RL--MPC controller introduced in \cite{yang2026resrl}. Here, the learned residual serves as the policy
transferred from simulation to the physical system.

At control step $k$, nominal MPC produces
$\mathbf u_k^{\rm mpc}\in\mathbb{R}^2$. A Soft Actor-Critic policy outputs
a normalized action $\mathbf a_k\in[-1,1]^2$, which is mapped to the
bounded residual velocity correction
\begin{equation}
\Delta\mathbf u_k=\alpha v_{\max}\mathbf a_k,
\label{eq:residual}
\end{equation}
where $\alpha$ determines the residual authority relative to $v_{\max}$.
The correction is activated only during detected pushing contact,
\begin{equation}
\tilde{\mathbf u}_k
=
\mathbf u_k^{\rm mpc}
+
I_{\rm ct}(k)\Delta\mathbf u_k,
\label{eq:resrl_raw}
\end{equation}
where $I_{\rm ct}(k)\in\{0,1\}$ is the contact indicator. The final command
is projected onto the admissible velocity set,
\begin{equation}
\mathbf u_k
=
\Pi_{\|\mathbf u\|_2\leq v_{\max}}
\!\left(\tilde{\mathbf u}_k\right),
\label{eq:resrl}
\end{equation}
so nominal MPC retains control during approach while the learned residual
acts during sustained pushing.

The policy is trained entirely in \micropush{} under time-varying flow and
task randomization. Prior simulation experiments showed improved robustness
and cross-geometry generalization relative to the nominal baselines
\cite{yang2026resrl}. For the experiments below, the trained policy is
frozen and used without real-world policy fine-tuning.

\subsection{Physical System and Evaluation Protocol}

The physical platform uses an approximately $20$~\um{} magnetic rolling
microrobot to manipulate silica particles of comparable size under
microscope feedback. Robot and particle positions are extracted online
from images, and the controller output is converted to the frequency and
heading of a rotating magnetic field.

We evaluate the transferred controller in two complementary tasks:
continuous figure-eight pushing and sequential multi-particle assembly.
The former tests sustained closed-loop manipulation over a long curved
trajectory with repeated contact transitions, while the latter evaluates
task-level integration of target assignment, global transfer planning,
and learned local pushing. The physical experiments were repeated across
multiple trials with successful executions observed repeatedly; the
results shown below were selected as representative demonstrations.

\begin{figure}
    \centering
    \includegraphics[width=\linewidth]{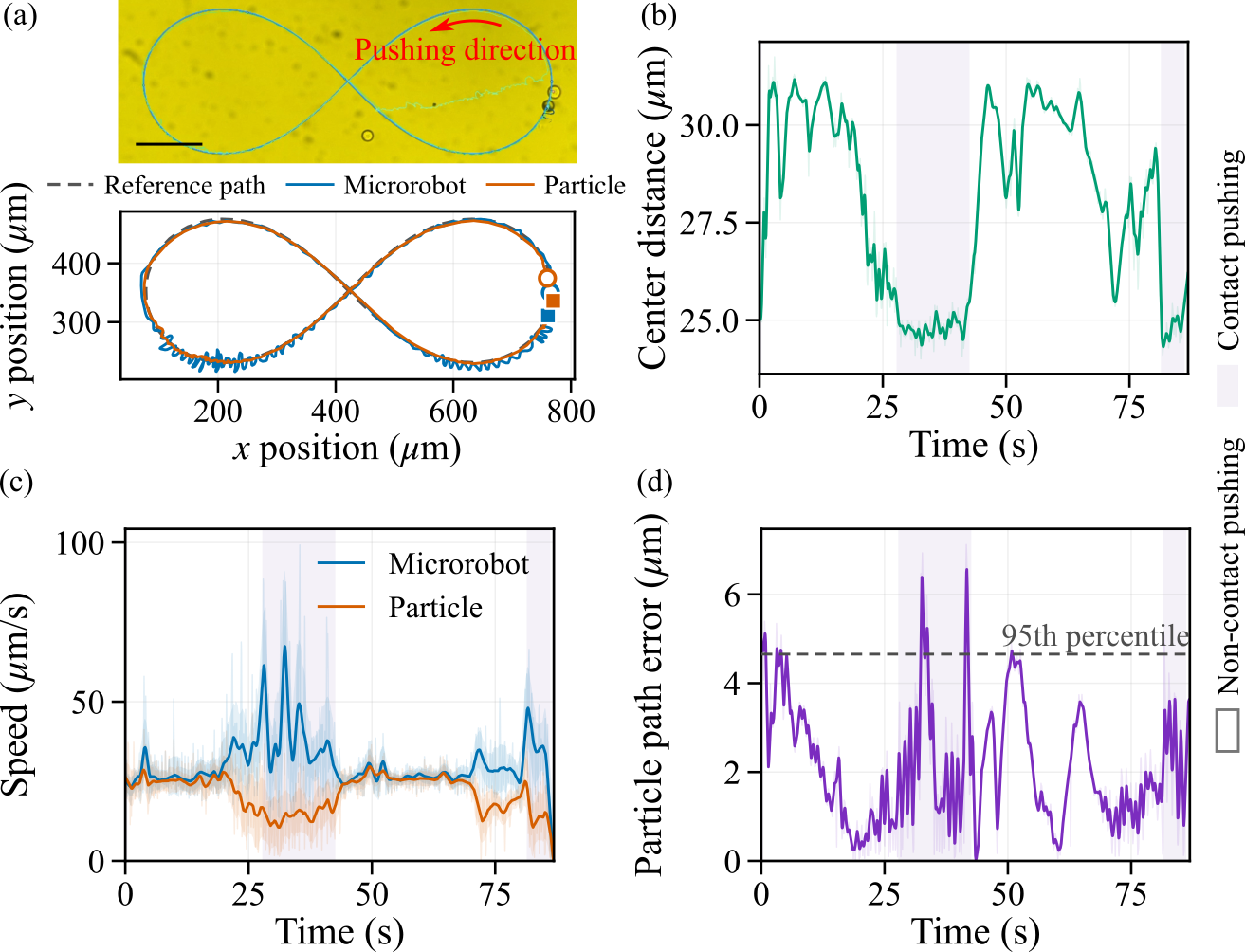}
    \caption{Zero-shot continuous pushing with the Residual RL--MPC policy trained in \micropush{}. (a) Microscope image and tracked microrobot/particle trajectories for the closed figure-eight task. (b) Robot--particle center distance; shaded intervals indicate the operational contact-pushing classification. (c) Microrobot and particle speeds. (d) Particle path error; the dashed line marks the 95th-percentile error of $4.65~\um$. The learned policy is deployed without real-world fine-tuning. The corresponding hardware experiment is shown in Movie S7.}
    \label{fig:figure8}
\end{figure}

\begin{figure*}
    \centering
    \includegraphics[width=0.8\textwidth]{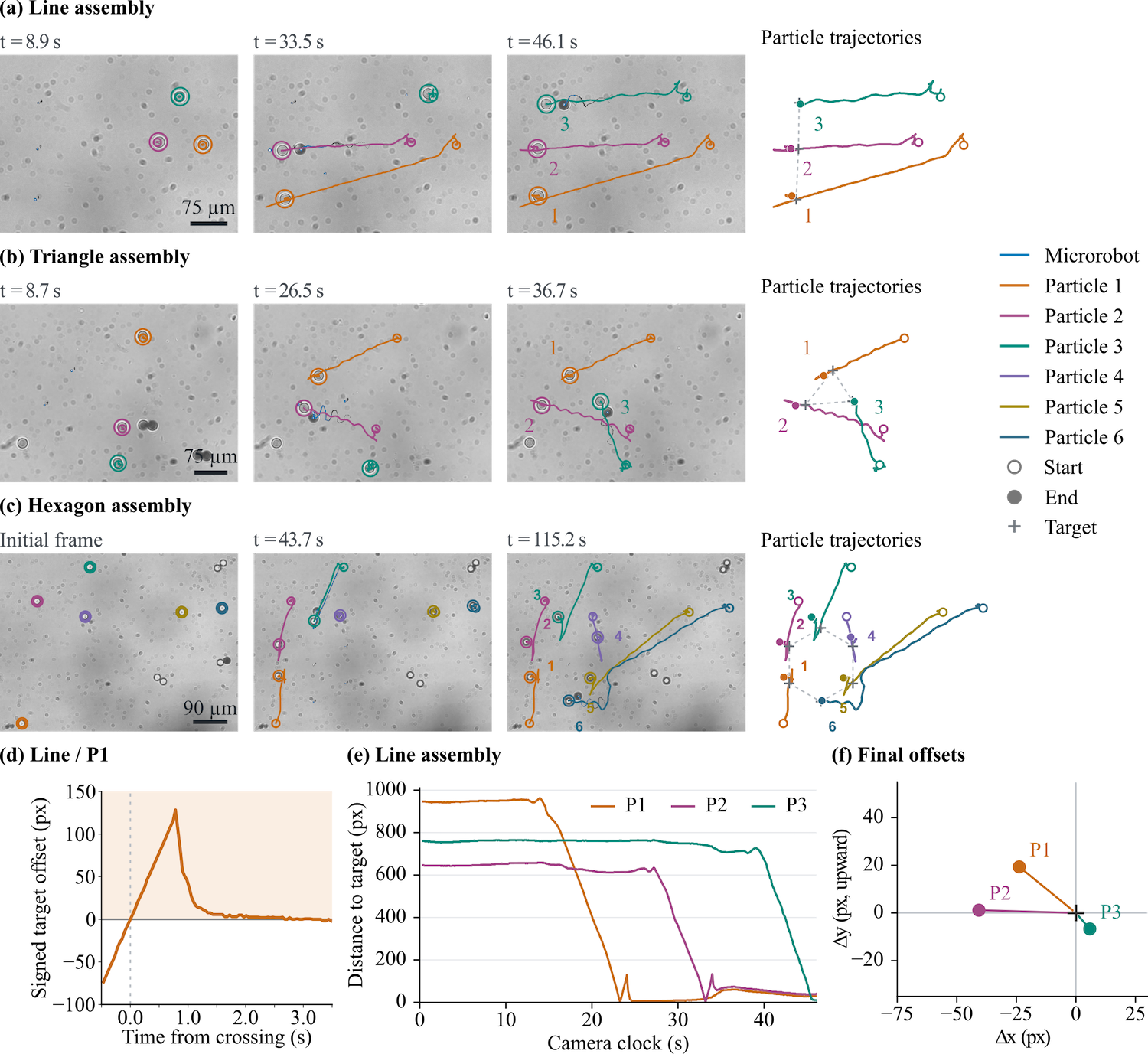}
    \caption{Sequential physical assembly and placement readouts using the same frozen Residual RL--MPC policy. (a--c) Representative line, triangle, and six-particle hexagonal assembly recordings. Each row shows selected microscope frames and the corresponding video-derived particle trajectories; markers denote particle starts, record-end positions, and target locations. (d) Signed target offset for particle P1 in the line assembly, aligned to its first crossing of the target plane; positive offset denotes intentional forward push beyond the target. (e) Particle-to-target distance over the line-assembly recording. (f) Two-dimensional record-end offsets relative to the line targets. Representative assembly experiments are shown in Movie S8.}
    \label{fig:assembly}
\end{figure*}

\subsection{Continuous figure-eight pushing}
Figure~\ref{fig:figure8} summarizes a continuous pushing experiment in which the particle follows a closed figure-eight trajectory. The retained trial lasts 86.89~s, during which the microrobot and particle travel 2686.02 and 1870.05~\um{}, respectively, corresponding to a cumulative transfer ratio of 0.696.

The robot--particle center distance in Fig.~\ref{fig:figure8}(b) is used to distinguish operational contact and non-contact pushing with a 25.5~\um{} threshold. This threshold is slightly larger than the nominal 20~\um{} center-to-center distance of two ideal 20~\um{} objects and is treated as an operational proxy that accommodates particle-size variation and tracking uncertainty. Contact pushing occupies approximately 20.15~s (23.2\%) of the trial. The remaining intervals primarily correspond to non-contact transfer and robot repositioning, during which short-range fluid-mediated coupling can contribute to particle motion.

The median microrobot and particle speeds are 27.84 and 23.49~\ums{}, respectively (Fig.~\ref{fig:figure8}(c)). Despite repeated changes in robot--particle configuration, the particle remains close to the prescribed trajectory: the median and mean path errors are 1.91 and 2.12~\um{}, the 95th-percentile error is 4.65~\um{}, and the maximum is 7.12~\um{} (Fig.~\ref{fig:figure8}(d)). The median robot--particle velocity-direction alignment is 0.906. The controller repeatedly re-establishes a useful pushing configuration after oscillation or temporary separation without manual intervention, demonstrating sustained closed-loop manipulation over a long curved trajectory rather than a single point-to-point push.

\subsection{Sequential multi-particle assembly}
Figure~\ref{fig:assembly}(a--c) extends the physical evaluation to sequential multi-particle manipulation. The same frozen Residual RL--MPC controller is integrated with task-level assignment and analytic geometry path planner \cite{yang2025microrobot}. Three particles are sequentially placed at user-specified targets for the line and triangle tasks. In the six-particle hexagon task, both target locations and particle--target assignments are generated automatically before the transfers are executed.

A practical physical effect appears during placement: after a particle reaches the target neighborhood, robot departure can displace it through near-field fluid coupling and surface interaction. We therefore use a small endpoint compensation,
\begin{equation}
\mathbf p_{\mathrm{push,end}}
=
\mathbf p_{\mathrm{target}}
+
d_{\mathrm{comp}}\hat{\mathbf t},
\label{eq:endpoint_comp}
\end{equation}
where $\hat{\mathbf t}$ is the terminal pushing direction and $d_{\mathrm{comp}}=9.4~\um{}$ in the reported experiments. The robot intentionally pushes slightly beyond the nominal target before disengagement so that the subsequent departure-induced motion can return the particle toward the desired location. This compensation is a physical-system adaptation outside the learned policy.

Figure~\ref{fig:assembly}(d) illustrates this mechanism for P1 in the line assembly: the signed offset becomes positive after the first target-plane crossing and subsequently returns toward zero. The zero-time alignment marks the first target-plane crossing. Figure~\ref{fig:assembly}(e) shows the three line-assembly particles approaching their respective targets sequentially, and Fig.~\ref{fig:assembly}(f) reports their two-dimensional offsets at the end of the recorded sequence.

The assembly experiments therefore test the simulator-trained local controller within a longer-horizon physical autonomy pipeline involving target generation, particle--target assignment, collision-aware planning, repeated contact acquisition, closed-loop pushing, and placement compensation. Successful line, triangle, and hexagonal assembly shows that the learned controller remains usable in a physical task-level loop despite effects such as substrate interaction, surface roughness, and departure-induced fluid coupling that are not represented exactly during training.

\section{Discussion}
The results demonstrate the practical value of \micropush{} as a task-level environment for microrobotic policy training and development. Despite its reduced-order interaction model, the zero-shot Residual RL–MPC experiments show that policies learned in \micropush{} can remain effective after transfer to a physical manipulation system. This is particularly relevant for microrobotics, where exhaustive hardware training is difficult and the practical value of a simulator depends not only on trajectory fidelity but also on whether it supports robust downstream autonomy.

\micropush{} currently uses a planar reduced-order interaction model that omits deformation, adhesion, three-dimensional flow, out-of-plane motion, and spatially varying magnetic fields. The physical experiments use silica particles; extending the framework to living cells will require additional validation of deformation, viability, and cell–substrate adhesion. The present hardware study demonstrates zero-shot transfer in continuous pushing and sequential assembly, while broader evaluation across robots, particle batches, substrates, and flow conditions remains for future work.

A natural next step is to preserve the speed and structure of \micropush{} while learning a residual dynamics model from synchronized physical trajectories. Such a model could use relative pose, gap, contact state, local image features, and tracking confidence to capture systematic effects omitted by the reduced-order simulator, including actuation bias, substrate interaction, and fluid-mediated coupling. Evaluation across held-out physical episodes could then quantify both predictive improvement and whether improved dynamics modeling translates into higher closed-loop success.

\section{Conclusion}
We presented \micropush{}, a lightweight simulator for microrobotic pushing and assembly that integrates reduced-order dynamics with planning, model-based control, RL, and reproducible evaluation. In addition to simulation benchmarks and faster-than-real-time execution, we evaluated \micropush{} through zero-shot policy transfer: a Residual RL--MPC policy trained entirely in simulation was deployed on the physical system without real-world policy fine-tuning, enabling accurate continuous figure-eight pushing and sequential line, triangle, and hexagonal assembly. These results support reduced-order simulation as a practical substrate for developing microrobotic autonomy while reserving physical experiments for validation and refinement rather than large-scale policy training.

\section{ACKNOWLEDGMENTS}
This work was supported by XXX.

\bibliographystyle{IEEEtran}
\bibliography{references}

\end{document}